# Blind signal separation and identification of mixtures of images

Felipe do Carmo, Joaquim T. de Assis, Vania Vieira Estrela, Alessandra M.Coelho
fpcarmo@iprj.uerj.br , joaquim@iprj.uerj.br, vestrela@iprj.uerj.br, amcoelho@iprj.uerj.br
Universidade do Estado do Rio de Janeiro (UERJ), Instituto Politécnico do Rio de Janeiro (IPRJ), CP 972825, CEP 28630-050, Nova Friburgo, RJ, Brazil

*Abstract*: In this paper, a fresh procedure to handle image mixtures by means of blind signal separation relying on a combination of second order and higher order statistics techniques are introduced. The problem of blind signal separation is reassigned to the wavelet domain. The key idea behind this method is that the image mixture can be decomposed into the sum of uncorrelated and/or independent sub-bands using wavelet transform. Initially, the observed image is pre-whitened in the space domain. Afterwards, an initial separation matrix is estimated from the second order statistics de-correlation model in the wavelet domain. Later, this matrix will be used as an initial separation matrix for the higher order statistics stage in order to find the best separation matrix. The suggested algorithm was tested using natural images. Experiments have confirmed that the use of the proposed process provides promising outcomes in identifying an image from noisy mixtures of images.

*Keywords: Blind source separation, independent component analysis, joint approximate diagonalization, performance index, image mixtures.*

## I. INTRODUCTION

In blind source separation (BSS), a mixture of observed signals $X(t)$ - also known as sensors - is supposed to be a function of independent sources $S(t)$. The main goal is to recover the source matrix $S(t)$ despite the lack of *a priori* information about sources and sensors the transfer function relating as well as the source signals $S(t)$. Better approaches to separate the sources present in a mixture can be found through statistical signal processing techniques combined with information theory, as for instance:

(i) *Joint-cumulants/cross-correlation zero forcing methods* are iterative procedures to solve simultaneous equations of joint cumulants [1,2] or cross-correlation [3]. In general, they are very inefficient and only work for static systems.

(ii) *Cross-correlation/energy minimization methods* result in adaptive rules found by means of a squared cross-correlation minimization [4] or an energy minimization [5]. These algorithms do not have sufficient constraints for source separation.

(iii) *Mutual information minimization (MIM) methods* output adaptive rules derived from mutual information minimization, either by [6] or [7] expansion, or with the help of a Parzen window [8], or from non-parametric quadratic mutual information [9]. These methods entail much higher computational complexity and are not advantageous to real-time implementation.

(iv) *Maximum a posteriori (MAP) methods estimation* requires that the convolution noise is modelled as a Gaussian distribution and Bussgang-type algorithms can be derived from the maximum likelihood [10, 11, 12].

(v) *Minimum entropy* **methods** **(MEM's)** are adaptive procedures derived empirically from the minimization of entropy [13].

As a rule, the performance of the methods from groups (iv) and (v) is poor, given that the probability density function (pdf) discrepancy is severe. Most of the solutions to BSS in this group consist of two steps [14. 15, 16].

This article focuses on *mutual information minimization* methods in the wavelet domain. Firstly, the separation matrix will be estimated by means of a second order statistics decorrelation method such as *joint-cumulants/cross-correlation zero forcing* because no tuning for the separation matrix is required. The observed data are linearly transformed, such that the correlation matrix of the output vector equals the identity matrix. Of course, when the initial separation matrix is selected near the true matrix, the rule will converge to the true separation matrix fast and the number of iterations will be reduced. This estimated matrix will be used as an initial guess for an iterative higher order statistics method that will compute the optimal (and final) estimate separation matrix.

During the second phase, the measured vector dimensionality can also be reduced to the same dimensionality of the source vector. After that, the separation matrix, between the whitened data and the output, will be an orthogonal matrix computed using an independent component (ICA) method.

Section II discusses the proposed algorithm. Section III presents some preliminary results for still images. Some conclusions are drawn in section IV.

## II. THE PROPOSED ALGORITHM

Let $S(k)$, and $X(k)$ be vectors constructed from the wavelet coefficients of the sources and mixtures of signals, respectively. The decomposition coefficients and the sources with linear mixing matrix are related by

$$X(k) = AS(k) \quad (1)$$

where $A$ is the mixing matrix. The relation between decomposition coefficients of the mixtures and the sources is exactly the same relation as in the original domain of signals, where $X(t) = AS(t)$. We shall estimate the mixing matrix $A$ or the separation matrix $B$ to build up the output estimated signals such $Y(t) = BX(t)$ using the wavelet coefficients of the mixtures $X(k)$ instead of $X(t)$. The wide-band input signal is whitened in time domain after the whitening signals used to estimate the decomposed into linear decomposition of several narrow-band subcomponents by applying the wavelet packet transform. Then, combination of the joint approximate diagonalization (JAD) algorithm and mutual information minimization in the wavelet domain will be used to solve the problem of BSS. The proposed algorithm can be divided into two main stages: initial estimation based on decorrelation approach, and natural gradient approach.

In the first step, the wavelet packet is used to decompose the wide-band input signal into several narrow-band components. Then, the joint approximate diagonalization algorithm (JAD) [14, 15, 16] is used to decorrelate the corresponding outputs of the wavelet transform to estimate the initial mixing matrix. JAD will be used to diagonalize the covariance matrices obtained from the corresponding components of the sub-band signals. It consists of two main steps: <u>Whitening</u> in the space domain [14] and <u>Separation</u> in the wavelet domain.

The pre-whitening phase operates on $X(f)$ whose eigenvectors are the orthogonal separating matrix. If vectors $S(f)$ are uncorrelated, then

$$E\{S_i(t)S_j^T(t)\} = 0 \quad \text{for } i \neq j \quad (2)$$

This situation is weaker than spatial independence, and it also can comprise nearly Gaussian-distributed sources. Besides this condition, suppose that each source signal has unit variance,

$$R_{SS}(0) = E\{S(t)S^T(t)\} = I \quad (3)$$

Consider the corresponding correlation matrix of the observed signals

$$\begin{aligned} R_{xx}(0) &= E\{X(t)X^T(t)\} \\ &= AE\{S(t)S^T(t)\}A^T \\ &= AR_{SS}(0)A^T \end{aligned} \quad (4)$$

The mixing matrix $A$ can be parameterized as $T D^2 G$ with $T$ and $G$ is unitary matrices and $D$ is a diagonal matrix. The correlation matrix of the observed signal is given by

$$R_{SS}(0) = AR_{SS}(0)A^T = AA^T = TD^2T^T \quad (5)$$

The eigenvalues decomposition of $\mathbf{R}_{xx}(0)$, which is unique, can be written as:

$$R_{xx}(0) = VCV^T \quad (6)$$

With $V$ is a unitary matrix and $C$ is a diagonal matrix. By identification, it is found that; $V = \mathbf{T}$ and $\mathbf{C} = \mathbf{D}^2$. The whitening matrix $W$ is then defined as $W = C^{-0.5}V^T$. The whitened signals are defined as:

$$Z(t) = WX(t) = WAS(t) = QS(t), \quad (7)$$

and the signal in the transform domain can be written as:

$$Z(k) = WX(k) = WAS(k) = QS(k) \quad (8)$$

The unitary matrix $Q$ must be estimated to estimate the initial separating matrix.

The wavelet packet output can be modelled as a linear combination of the sub-band components, defined as:

$$Z(k) = Z_1(k) + Z_2(k) + ... + Z_m(k) \quad (9)$$

$$= Q(S_1(k)+S_2(k)+...+S_m(k)),$$

where $m$ is the number of sub-bands. The covariance matrix of the transformed whitened signals can be written as:

$$R_{zzi}(0) = WAR_{ssi}A^TW^T = QR_{ssi}(0)Q^T \quad (10)$$
$$1 < i \leq m$$

Now from each corresponding sub-band, the covariance matrix is estimated to construct a set of covariance matrices. Then, the joint approximate diagonalization (JAD) process, which is an iterative technique of optimization over the set of orthonormal matrices, is obtained as a sequence of plane rotations. The objective of this procedure is to find the orthogonal matrix $Q$ which diagonalizes a set of matrices applied on this set to obtain on the unitary matrix $Q$, and the initial separating matrix will be

$$B_{initial} = Q^TW. \quad (11)$$

A cost function accounting for all higher-order statistics is introduced in [17]. Their primary novelty was to use mutual information, as a measure of statistical independence. The mutual information $I(Y_1,Y_2)$ measures the degree of overlap between two random variables, $Y_1$ and $Y_2$. It is always positive, and zero if and only if $Y_1$ and $Y_2$ are independent [18]. Mathematically, the independence of $Y_1$ and $Y_2$ can be expressed by the relationship:

$$p(Y_1,Y_2) = p(Y_1)p(Y_2), \quad (12)$$

where $p(Y_1,Y_2)$ is the joint probability density function of two variables $Y_1$ and $Y_2$ and $p(Y_1)$, $p(Y_2)$ is the probability density function of each variable $Y_1$, $Y_2$ respectively. Writing the mutual information among $n$ variables in terms of a more quantifiable measure, the joint entropy $H(y_1,...,y_n)$, can be rewritten via the chain rule:

$$H(y_1,...,y_n) = H(y_1)+...+H(y_n)-I(y_1,...,y_n). \quad (13)$$

Implying that

$$H(y_1,...,y_n) = -E\{\log p(y_1)\}-...+...$$
$$+E\{\log p(y_2)\}-I(y_1,...,y_n) \quad (14)$$

The above equation demonstrates clearly that simply maximizing the joint entropy of the outputs $Y(t)$ is not the same as minimizing the mutual information, due to the interfering marginal entropy terms. However, if $Y(t) = g(u)$, where $g$ is an invertible function so that (by simple Jacobin transformation),

$$p(y_i) = \frac{p(u_i)}{\left|\frac{\partial g(u_i)}{\partial u_i}\right|} \quad (15)$$

Then the marginal terms can be eliminated by setting

$$g(u_i) = p(u_i) = \frac{\partial y_i}{\partial u_i}. \quad (16)$$

In this case, we have

$$H(y_1,...,y_n) = -I(y_1,...,y_n) \quad (17)$$

This implies that the nonlinearity $g(u)$ has the form of the cumulative density function of the true source distribution. As a consequence, maximization of the joint entropy of $Y(t)$ is equivalent to minimizing the mutual information between the components of $Y(t)$.

If a suitable $g(u)$ can be found, so that the marginal error terms are negligible, it is possible to obtain a cost function depending on the information content:

$$J = H(Y) = H(g(u)) = -E\left[\log p\left(g\left(B(t)Z(k)\right)\right)\right] \quad (18)$$

It is clear that $\frac{\partial J}{\partial B(t)}$ gives a *deterministic* gradient ascent direction to establish the maximum. Due to the expectation operator, this involves block estimation of averages over $Z(k)$. An alternative strategy is to take out the expectation operator, thus using stochastic gradient. This gradient is perturbed by the local random motion of $Z(k)$, but still eventually converges, given the average effect on search directions on a global scale. Stochastic gradient methods benefit from having better tracking aptitude. Then, the final objective function becomes

$$(14) \quad J = \log p\left(g\left(B(t)Z(k)\right)\right). \quad (19)$$

By computing $\dfrac{\partial J}{\partial B(t)}$ then:

$$\Delta B \alpha \dfrac{\partial J}{\partial B} = (B^T)^{-1} + \dfrac{g''(u)}{g'(u)} Z_i^T(k). \qquad (20)$$

Accordingly, the gradient ascent update can be given by:

$$B(i+1) = B(i) + \mu[(B^T(i))^{-1} + \dfrac{g''(u)}{g'(u)} Z_i^T(k)] \qquad (21)$$

where $\mu$ is the step size or learning rate and $B(0) = B_{initial}$. The estimated separation will be the final estimation $B_{final}$. If the non-linearity is taken as $\varphi(u) = \dfrac{g''(u)}{g'(u)}$:

$$B(i+1) = B(i) + \mu\left[\left(B^T(i)\right)^{-1} + \varphi(u) Z_i^T(k)\right]. \qquad (22)$$

A much more efficient search direction can be obtained by post-multiplying the entropy gradient in (17) by $B^T B$ [19]:

$$\begin{aligned} B(i+1) &= B(i) + \\ & \mu\left[\left(B^T(i)\right)^{-1} + \varphi(u) Z_i^T(k)\right] B^T(i) B(i) \\ &= B(i) + \mu\left[I + \varphi(u) Z_i^T(k)\right] B(i), \end{aligned} \qquad (23)$$

which leads to the so-called *natural gradient* algorithm. It is clear from the standard gradient in equation (20) that the convergence depends on the axis scaling [20], while the natural gradient algorithm normalizes $B$, rendering the gradient invariant to such scaling. In the proposed algorithm, $g(u) = \tanh(u)$ is used as nonlinear function with learning rate $\mu = 0.00002$. The proposed method can be summarized in the following algorithm:

**Algorithm:**

```
(1) Apply the whitening matrix W to the
source signal X(t).
(2) Perform the Wavelet Transform.
(3) Decorrelate the result from step 1
with JAD.
(4) Calculate an initial estimation for
the separation matrix.
(5) Apply the natural gradient method to
the results of steps 2 and 4.
```

## III. EXPERIMENTS

Two natural still images were used to evaluate the performance of the proposed technique and to confirm its effectiveness with and without noise. Later, the technique was used to separate images for the cases of: white noise, impulse noise and the mixture of them. A fine metric for rating the separation quality is the performance index (*PI*) [21], which is defined as:

$$PI = \dfrac{1}{n(n-1)} \sum_{i=1}^{n} \sum_{j=1}^{n} \left( \dfrac{[B]_{ij}}{\max_j [B]_{ij}} - 1 \right), \qquad (24)$$

where $[B]_{ij}$ is the ($i,j$)-element of the matrix $B$, and $n$ is the number of sources. Generally, increasing *PI* results in a performance improvement.

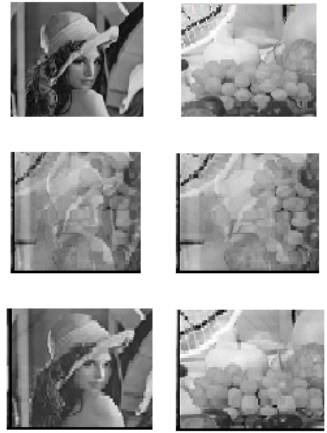

Figure 1. Results of separating mixtures of two noise-free images.

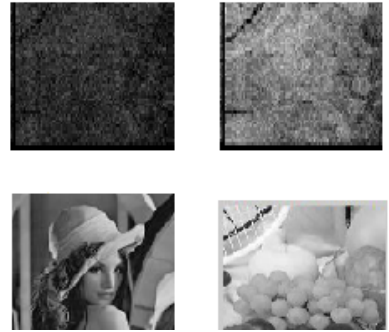

Figure 2. (a) Results for the mixed images of the earlier figure with additive white noise (SNR=15 dB); and (b) output of the separation procedure.

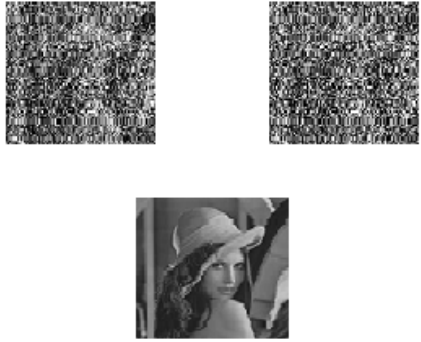

Figure 3. (a) The original corrupted images by white Gaussian noise with SNR= -8 dB and Salt & Pepper noise; and (b) the estimation result.

The comments in this paragraph are related to Fig. 1, which contains two natural images (Lena and Fruits) for the noise-free case. The original frames are shown in its upper part. Their mixtures appear in middle: the left frame with 'Lena" as the prevailing source, corrupted by "Fruits"; and the right frame contains "Fruits" corrupted by "Lena". The estimation results of the algorithm are presented in the bottom.

Then, the mixed images of the previous experiment were blurred by additive white noise with *SNR=15* dB as can be seen from Fig. 2a. The results obtained by means of the proposed methodology - that is, via separating matrices - appear in Fig. 2b.

Fig. 3a demonstrates the algorithm performance when "Lena" is corrupted by white Gaussian noise with SNR= -8 dB and Salt & Pepper noise with noise density of 40%. *PI*'s of 0.0213 and 0.0049 were achieved in the decorrelation step, after the second stage.

## IV. CONCLUSIONS

A new procedure to solve the problem of blind source separation for the case of linear mixtures of images was introduced. The novel algorithm combines two stages to combine the advantages of each one. The proposed procedure is based on minimizing the zero time lag cross-correlation between the decomposed corresponding components of the wavelet packet transformation outputs followed by minimizing the mutual information process. The estimated separating matrix from the second order technique is used as initial separation matrix with the higher order statistic (natural gradient). From the simulation results, the iterative process converged fast with optimum separating matrix. The algorithm was tested using different natural images to evaluate its performance. Experiments have confirmed that the use of the proposed procedure provides promising results. Moreover, this technique succeeds in sorting out images from noisy mixtures.

ACKNOWLEDGEMENTS

The authors are thankful to CAPES, FAPERJ and CNPq for scholarships and grants received.

REFERENCES

[1]  D. Yellin and E. Weinstein, "Criteria for multichannel signal separation", IEEE Trans. on Signal Processing, vol. 42, no. 8, pp. 2158-2168, August 1994.

[2]  D. Yellin and E. Weinstein, "Multichannel signal separation: methods and analysis," IEEE Trans. on Signal Proc., vol. 44, no. 1, pp. 106-118, January 1996.

[3]  E.Weinstein, M. Feder and A. V. Oppenheim, "Multi-channel signal separation by decorrelation", IEEE Trans. on Speech and Audio Processing, vol. 1, no. 4, pp.405-413, October 1993.

[4]  M. Najar, M. Lagunas, and I. Bonet, "Blind wideband source separation", in Proc. 1994 IEEE Int. Conf. on Acoustics, Speech and Signal Proc., Adelaide, South Australia, pp. 65-68, April 1994.

[5]  S. V. Gerven and D. Van Compernolle, "Signal separation by symmetric adaptive decorrelation: stability, convergence, and uniqueness", IEEE Trans. on Signal Processing, vol. 43, no. 7, pp. 1602-1611, July 1995.

[6] P. Comon, "Independent component analysis: A new concept", Signal Processing, vol. 36, no. 3, pp. 287-314, 1994

[7] H. H. Yang and S.-I. Amari, "Adaptive online learning algorithms for blind separation: maximum entropy and minimum mutual information", Neural Computation, vol. 9, no. 7, pp. 1457-1482, October 1997

[8]  D. T. Pham, "Blind separation of instantaneous mixture of sources via an independent component analysis", IEEE Transactions on Signal Processing, vol. 44, no. 11, pp. 2768-2779, November 1996.

[9]  D. Xu, Energy, Entropy and Information Potential for Neural Computation, Ph.D. dissertation, University of Florida, Gainesville, Florida, USA, 1999.

[10]  S. Bellini, "Bussgang techniques for blind deconvolution and equalization" , Blind deconvolution, pp. 8-52, Englewood Cliffs, NJ, Prentice Hall, 1994.

[11]  S. Haykin, Blind Deconvolution, Englewood Cliffs, NJ, Prentice Hall, 1994.

[12]  R. H. Lambert, "A new method for source separation," in Proc. of IEEE Int'l Conf. on Acoustics, Speech and Sig. Proc., vol. 3, pp. 2116-2119, 1995.

[13]  A.T. Walden, "Non-Gaussian reflectivity, entropy and deconvolution", Geophysics, vol. 50, no. 12, December 1985, pp. 2862-2888.

[14]  A. Belouchrani, K. Abed-Meraim, J. Cardoso, and E. Moulines, "A blind source separation technique using second-order statistics," IEEE Trans. Signal Processing, vol. 45, pp. 434 - 444, Feb. 1997.

[15]  J. Cardoso, and A. Souloumiac, "Blind beamforming for non-Gaussian signals," Proc. IEE- F, vol.140, no.6, pp.362-370, Dec.1993.

[16] A. Belouchrani, and M. G. Amin, "Blind source separation based on time-frequency signal representation," IEEE Trans. Signal Proc., vol. 46, no.11, pp 2888-2898, Nov 1998.

[17] A. Bell, and T. Sejnowski, "An inmation maximization approach to blind separation and blind deconvolution," Neural Computation, 11: 157–192 1999.

[18] T.Cover, J.Thomas, Elements of Information Theory. John Wiley & Sons, 1991.

[19] S. Amari, "Natural Gradient Works Efficiently in Learning. Neural Computation, 10:251-276, 1998.

[20] T. W. Lee," Independent component analysis: theory and applications," Boston: Kluwer Academic Publisher, 1998.

[21] A. Cichocki, and S. Amari, "Adaptive blind signal and image processing: learning algorithms and applications," England: John Wiley & Sons, Chichester, 2002.

[21] C. Choi, A. Cichocki, and S. Amari, "Flexible independent component analysis", Journal of VLSI Signal Processing-Systems for Signal, Image, and Video Technology, 2000.

[22] K. Blekas, A. Likas, N.P. Galatsanos, I.E. Lagaris, "A spatially-constrained mixture model for image segmentation", IEEE Trans. on Neural Networks,vol. 16, 494-498, 2005.

[23] A.K. Katsaggelos and N.P. Galatsanos, Signal recovery techniques for image and video compression and transmission, Springer, 1998